\documentclass{article}

\usepackage{arxiv}

\usepackage[utf8]{inputenc} 
\usepackage[T1]{fontenc}    
\usepackage{hyperref}       
\usepackage{url}            
\usepackage{booktabs}       
\usepackage{amsfonts}       
\usepackage{amsmath}
\numberwithin{table}{section}
\usepackage[frozencache,cachedir=.]{minted}
\usepackage{pifont}
\newcommand{\cmark}{\Pisymbol{pzd}{51}}%
\newcommand{\xmark}{\Pisymbol{pzd}{55}}%

\usepackage{nicefrac}       
\usepackage{microtype}      
\usepackage{lipsum}

\usepackage{caption} 
\usepackage{makecell}
\usepackage{multirow}
\usepackage{hhline}
\usepackage{rotating}
\usepackage[flushleft]{threeparttable}

\usepackage{amsmath}
\usepackage{amssymb}

\numberwithin{equation}{section}
\numberwithin{figure}{section}

\usepackage{url} 
\usepackage[table]{xcolor}
\definecolor{dgreen}{rgb}{0.0, 0.5, 0.0}
\definecolor{dred}{rgb}{0.8, 0.0, 0.0}
\usepackage{float}

\usepackage{listings}

\usepackage{array, multirow,graphicx}
\usepackage{adjustbox}

\usepackage{placeins}
\usepackage{hhline}
\let\Oldsection\section
\renewcommand{\section}{\FloatBarrier\Oldsection}

\usepackage{natbib}
\title{Multi-Agent Environments for Vehicle Routing Problems}

\author{
Ricardo Gama\\
School of Technology and Management of Lamego\\
Polytechnic Institute of Viseu, Portugal\\
\texttt{rgama@estgl.ipv.pt} \\
\And
Ricardo Cunha\\
School of Technology and Management of Lamego\\
Polytechnic Institute of Viseu, Portugal\\
\And
Daniel Fuertes\\
Grupo de Tratamiento de Imágenes (GTI),\\ 
Information Processing and Telecommunications Center, ETSI \\ Telecomunicación, Universidad Politécnica de Madrid\\
\texttt{d.fcoiras@upm.es} \\
\And
Carlos R. del-Blanco\\
Grupo de Tratamiento de Imágenes (GTI), \\
Information  Processing and Telecommunications Center, ETSI\\ 
Telecomunicación, Universidad Politécnica de Madrid\\
\texttt{carlosrob.delblanco@upm.es} \\
   \And
 Hugo L. Fernandes \\
  Singuli\\
  New York, USA\\
  \texttt{hugoguh@gmail.com} \\
}

\begin{document}
\maketitle

\begin{abstract}
Research on Reinforcement Learning (RL) approaches for discrete optimization problems has increased considerably, extending RL to areas classically dominated by Operations Research (OR). Vehicle routing problems are a good example of discrete optimization problems with high practical relevance, for which RL techniques have achieved notable success. Despite these advances, open-source development frameworks remain scarce, hindering both algorithm testing and objective comparison of results. This situation ultimately slows down progress in the field and limits the exchange of ideas between the RL and OR communities. Here, we propose MAEnvs4VRP library, a unified framework for multi-agent vehicle routing environments that supports classical, dynamic, stochastic, and multi-task problem variants within a single modular design. The library, built on PyTorch, provides a flexible and modular architecture design that facilitates customization and the incorporation of new routing problems. It follows the Agent Environment Cycle ("AEC") games model and features an intuitive API, enabling rapid adoption and seamless integration into existing reinforcement learning frameworks. The project source code can be found at \url{https://github.com/ricgama/maenvs4vrp}. %
\end{abstract}

\section{Introduction}

Since the seminal work on Pointer network models (\cite{Vinyals2015, Bello2016}), there has been significant growth in machine learning approaches to solving discrete optimization problems. These pioneering studies outlined two distinct approaches for training Pointer networks to solve discrete optimization problems: supervised learning (\cite{Vinyals2015}), where the network is trained on existing solutions to the problem, and reinforcement learning (\cite{Bello2016}), where the network learns by evaluating the quality of the solutions it generates while attempting to solve the problem. While a supervised learning approach may be desirable when ground-truth data is readily available, it becomes a limitation for some combinatorial optimization problems where data collection is infeasible. Moreover, its performance is constrained by the quality of the existing solutions. In contrast, a reinforcement learning approach bypasses the need for ground-truth data in model training, thereby extending the applicability of Pointer networks models to a larger set of practical discrete optimization problems (see \cite{MAZYAVKINA2021105400} for a recent overview).

One type of application where development has been particularly notable is Vehicle Routing Problems (VRPs). The VRPs are a generic class of optimization problems focused on determining the optimal route for a fleet of vehicles tasked with serving a collection of customers, while satisfying specific constraints (\cite{braekers2016vehicle}). Following the seminal works of \cite{Nazari2018} and \cite{Kool2019a}, which use Pointer network models to solve VRPs, the field has witnessed substantial growth, both in the development of new methods and in the extension to other VRP variants (see \cite{Lireview2022, SHI2023773, wu2024neural, zhou2024learning} for recent reviews). Furthermore, recent research (\cite{liu2024multi, zhou2024mvmoe, berto2024routefinder}) has proposed more general and versatile models, capable of concurrently addressing various VRP. These advancements facilitate the development of high-performance solvers that can generalize across different problem specifications.

While these achievements are significant, many existing approaches to solving VRPs simplify the complexity of multi-vehicle fleet problems by modeling them as single-agent (vehicle) problems, where a vehicle repeatedly returns to its depot. This reduction overlooks the problem's multi-agent nature and may limit the benefits of cooperative learning, thereby affecting decision-making and solution quality. Recent research suggests that multi-agent architectures can achieve superior performance compared to single-agent counterparts in specific deterministic environments (e.g., \cite{berto2024parco}). Moreover, while single-agent approaches may yield reasonable solutions for offline planning, they are not suitable for online applications in which fleet vehicles must make instantaneous decisions based on the current state of the environment. This is particularly relevant in practical applications where unpredictability is present and prompt decision-making is needed. Therefore, adopting a multi-agent strategy can offer significant advantages across deterministic, dynamic, and stochastic routing problems, fostering the development of resilient, adaptable systems that meet real-world operational demands. Some works tackling these aspects have appeared, proposing multi-agent approaches for dealing with several variations of routing problems, such as capacity vehicle routing (\cite{bono2020solving}, \cite{zhang2020multi}, \cite{zhang2023coordinated}, \cite{arishi2023multi}, \cite{arishi2023multi}, \cite{berto2024parco}, \cite{liu20242d}), orienteering (\cite{fuertes2023}), pickup and delivery (\cite{zong2022mapdp}, \cite{10417723}), and dynamic vehicle routing problems (\cite{GUO2023103095, pan2023deep}). 

To successfully develop multi-agent models/solutions for discrete optimization problems, it is necessary to have access to common development frameworks and simulation environments. However, these resources remain limited, particularly for VRPs, hindering objective comparisons, the exchange of ideas, and the overall development of the field. To address this limitation, we propose a library built on PyTorch (\cite{pytorch}) that provides multi-agent environments for simulating classical vehicle routing problems and lays down a framework for further generalizations.  We primarily follow the logic of the \textit{PettingZoo} API and the design principles of the \textit{Flatland} environments library (\cite{Flatland}). Adopting these principles improves the generality and usability of our environments. Our library offers a flexible, modular architecture that enables easy customization and implementation of new VRPs. It features a user-friendly API for rapid adoption and smooth integration with existing reinforcement learning frameworks. The library includes benchmark instance sets for each environment, as well as baseline neural network models and training code.

VRPs encompass a diverse range of problems and are designed to address increasingly complex real-world applications. As a starting point, we chose to implement environments mainly centered on VRPs with time windows. This class covers a broad spectrum of practically significant problems and can serve as a robust foundation for future generalization to other, more complex scenarios.

The main contributions of the proposed Multi-agent Environments for Vehicle Routing Problems library (MAEnvs4VRP) are:
\begin{itemize}
\item A general and modular design that enables easy implementation and generalization to other routing problems.
\item A seamless adoption of multi-agent environments that supports both online and offline applications.
\item A familiar API structure, which enables integration with existing multi-agent reinforcement learning algorithm training and development platforms.
\item A simple integration of benchmark instances and easy implementation of new instances, ensuring clean and reproducible definitions of train, validation, and testing sets.
\end{itemize}

This manuscript is organized as follows. Section \ref{sec:related_work} provides a first overview of related work. Section \ref{sec:design} describes the library’s architecture, including its primary structure and core functionalities. Section \ref{sec:experiments} presents a set of performance experiments and baselines applicable to a subset of the available environments. Section \ref{sec:disc} discusses the current development status and outlines future directions.

\section{Background and Related Work}
\label{sec:related_work}

The recent success and rapid growth of reinforcement learning (RL), particularly in the domain of multi-agent reinforcement learning (MARL), have been accompanied by and promoted through the emergence and establishment of several development frameworks (e.g., \cite{raffin2021stable, bou2023torchrl, hu2023marllib}). Libraries based on standardized APIs, such as Gym/Gymnasium  (\cite{brockman2016openai, Gymnasium2024}) and PettingZoo (\cite{PettingZoo}), allow the development, testing, and comparison of algorithms on a common platform. In fact, as the field continues to expand, the need for standardization and reproducibility becomes a crucial concern, not only within specific research communities (like MARL, \cite{bettini2024benchmarl}) but also across intersecting disciplines, such as the RL and OR communities in the context of combinatorial optimization (\cite{Accorsi20222}).

\subsubsection*{Existing libraries with VRP environments.} 

Although still scarce, recent years have witnessed the development of several libraries offering RL environments designed for specific discrete optimization problems. Particularly \textit{ORL} (\cite{balaji2019orl}), \textit{OR-Gym} (\cite{hubbs2020orgym}), Graphenv  (\cite{biagioni2022graphenv}), \textit{RLOR} (\cite{wan2023rlor}) and \textit{Jumanji} (\cite{bonnet2023jumanji}) provide a suite of environments for various operations research problems, such as Knapsack, Bin Packing, Inventory and Network Management, Vehicle Routing and Traveling Salesman.

\begin{table}[htbp]
\centering
\begin{adjustbox}{width=1\textwidth}
\begin{tabular}{lccccccc}
\toprule
Library           & \multicolumn{1}{l}{\begin{tabular}[c]{@{}l@{}} \# VRP\\ enviroments\end{tabular}} & \multicolumn{1}{c}{Multi-agent} & \multicolumn{1}{c}{Vectorized} & \multicolumn{1}{c}{\begin{tabular}[c]{@{}c@{}}Customizable\\   generation\end{tabular}} & \multicolumn{1}{c}{\begin{tabular}[c]{@{}c@{}}Customizable\\  observations\end{tabular}}& \multicolumn{1}{c}{\begin{tabular}[c]{@{}c@{}}Customizable\\  rewards\end{tabular}} & \multicolumn{1}{c}{\begin{tabular}[c]{@{}c@{}}On\slash Offline\end{tabular}} \\ \hhline{========}
ORL (\cite{balaji2019orl})            & 1                                      & \xmark                              & \xmark                              & \xmark                                                                                                    & \xmark & \xmark                                                                                             & online                                \\
OR-Gym (\cite{hubbs2020orgym})           & 2                                      & \xmark                              & \xmark                              & \xmark                                                                                                    & \xmark & \xmark                                                                                             & offline                               \\
Graphenv (\cite{biagioni2022graphenv})        & 1                                      & \xmark                              & \xmark                              & \xmark                                                                                                    & \xmark & \xmark                                                                                             & offline                               \\
RLOR (\cite{wan2023rlor})           & 2                                      & \xmark                              & \cmark                              & \xmark                                                                                                    & \xmark & \xmark                                                                                             & offline                               \\
RoutingArena  (\cite{thyssens2023routing})    & 1                                      & \xmark                              & \cmark                              & \cmark                                                                                                   & \xmark & \xmark                                                                                             & offline                               \\
Jumanji   (\cite{bonnet2023jumanji})        & 3                                      & \cmark ($^\ddagger$)                          & \cmark                              & \cmark                                                                                                   & \xmark & \xmark                                                                                             & offline                               \\
RL4CO  (\cite{berto2023rl4co})        & 20                                     & \xmark                              & \cmark                              & \cmark                                                                                                    & \xmark & \xmark                                                                                             & offline                               \\ \midrule
Maenvs4VRP (ours) & 13                                      & \cmark                              & \cmark                              & \cmark                                                                                                   & \cmark & \cmark                                                                                             & both                                  \\ \bottomrule
\multicolumn{8}{l}{\footnotesize{$\ddagger$ It has one multi-agent environment}} \\
\end{tabular}

\end{adjustbox}
\caption{Comparison of existing libraries that provide VRPs environments}
\label{table:libs}
\end{table}

In a broader context, two general frameworks, \textit{RoutingArena} (\cite{thyssens2023routing}) and \textit{RL4CO} (\cite{berto2023rl4co}), have been developed to facilitate reproducible research and algorithm benchmarking. \textit{RoutingArena} focuses on the Capacitated VRP and offers a benchmark platform that includes popular OR baselines and a wide range of benchmark instances. It enables the comparison of different RL and OR solvers using standardized evaluation metrics. The \textit{RL4CO} library provides a unified framework that integrates environments, policies, and reinforcement learning algorithms into a single comprehensive package. In addition, several isolated environments are also available as accompanying material for research papers (e.g., \cite{Kool2019a, kwon2020pomo, kim2022sym, zhang2023first, berto2024parco}). These environments are typically limited to the specific problems for which they were developed, and this limitation can significantly hinder their adaptability or reusability in other contexts.

While these tools represent significant progress towards creating unified frameworks for research and development, they focus almost exclusively on single-agent environments (see Table \ref{table:libs}). This emphasis overlooks elements that could be effectively addressed using multi-agent reinforcement learning techniques, such as collaboration among fleet vehicles. Additionally, they often lack the flexibility to customize different functional components of the environments that are particularly relevant for RL, such as rewards and observations. This limitation restricts the exploration of new ideas and the ability to customize environments to suit different application constraints. To address these issues, we developed \textit{MAEnvs4VRP} from the ground up, utilizing a modular design that enables the independent customization of the various environment components. By decoupling these elements, the library provides a platform to evaluate how specific configurations of observability and reward structures influence agent coordination.

\subsubsection*{Sequential Decision-Making in Multi-Agent VRP.} 

The design of reinforcement learning environments and APIs, particularly those involving MARL, is closely linked to the formal game models they adopt. Most MARL research commonly relies on one of two underlying formal game models: Partial Observable Stochastic Games (\cite{marl-book}) and Extensive Form Games (\cite{shoham2008multiagent}). While widely used, these models suffer from some drawbacks when transitioning from the abstract model to its practical code implementation (\cite{PettingZoo}). In an attempt to mitigate this weakness, PettingZoo (\cite{PettingZoo}) introduced the Agent Environment Cycle (AEC) games model. This model offers a conceptual alternative to commonly used models, providing advantages in the practical implementation of multi-agent environments and their API design. Specifically, in the AEC model, each agent acts sequentially, seeking to resolve 'tie-breaking' decisions (conflict handling) that often occur in multi-agent settings, where agents are allowed to choose actions simultaneously. In addition, sequential acting also allows for clear information management in environments where the number of agents can change throughout an episode (rollout) due to creation or elimination processes. Another significant feature is that the AEC model enables rewards to be distributed either collectively or individually during or after the episode, providing flexibility for exploring reward engineering strategies.

\begin{figure}[ht]
\includegraphics[width=0.75\textwidth]{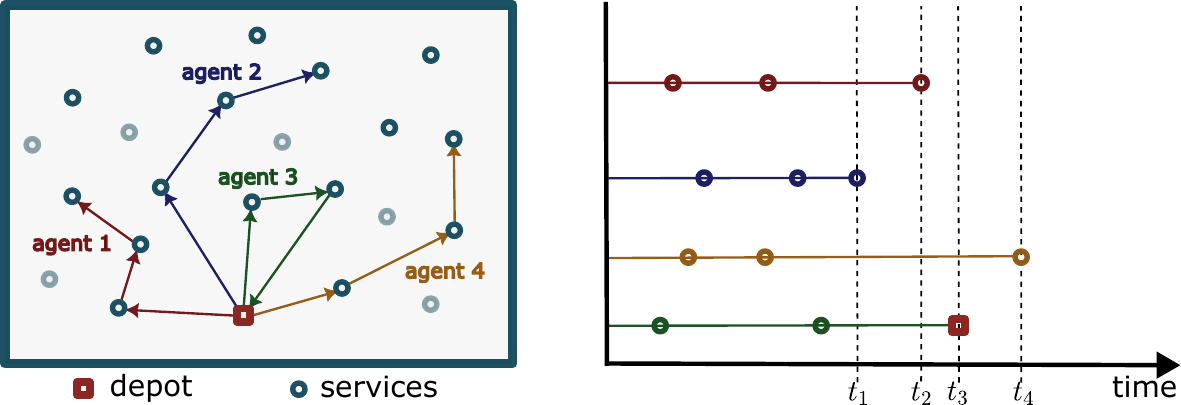}
\centering
\caption{Illustration of a multi-agent VRP instance with four vehicles (left) and the corresponding timeline (right). All vehicles have completed a series of three action steps. Since agents act asynchronously, the next agent to interact with the environment might have only partial information available.}
\label{inst_fig}
\end{figure}

These characteristics of the AEC game model are particularly crucial in VRP, and similar ideas were explored in \cite{bono2020solving}, where the authors introduced a sequential multi-agent Markov decision process to address multi-agent dynamic vehicle routing problems. For time-dependent VRPs, each agent's (vehicle's) action is extended over time, and as a consequence, agents must make asynchronous decisions. If agents acted simultaneously, the relative timing of the events would be lost, limiting the applicability of learning policies in these environments to real-world scenarios (\cite{menda2018deep}). Furthermore, the various constraints inherent to routing problems require conflict resolution rules.

As an example, consider the problem illustrated in Figure \ref{inst_fig}. Four agents have already completed three steps in an environment where each service can only be performed by a single vehicle. If agents act simultaneously during a specific step of a given environment, the environment must internally handle conflicting actions and apply tiebreaker rules to determine which action each agent should execute. This approach has two main drawbacks. First, once an agent’s action is selected, the environment’s state changes, causing the other agents to base their decisions on outdated observations (\cite{PettingZoo}). Second, the assumption of simultaneous action by all agents does not directly lead to practical applications in which agents must decide their actions at different times (c.f. \cite{menda2018deep}). These issues are mitigated when agents operate sequentially, as decisions are made based on the latest environmental observations and can account for the timelines of other agents. These observations led us to adopt the AEC model as one of the foundations of our library.

\section{The MAEnvs4VRP Library Architecture\label{sec:design}}

This section details the MAEnvs4VRP library architecture and API design. Our core philosophy is to implement each environment independently while enforcing a uniform high-level structure across all variants. This balance between independence and uniformity promotes clarity, maintainability, and extensibility across the framework. MAEnvs4VRP explicitly separates architectural abstractions from problem-specific routing logic. This separation enables users to easily understand the implementation details of individual environments and facilitates customization, extension, and generalization to new problem settings. As a result, the library is designed to function both as a flexible research platform and as a standardized benchmarking toolkit.

\subsection{Core Components}
 
\begin{figure}[ht]
\includegraphics[width=0.50\textwidth]{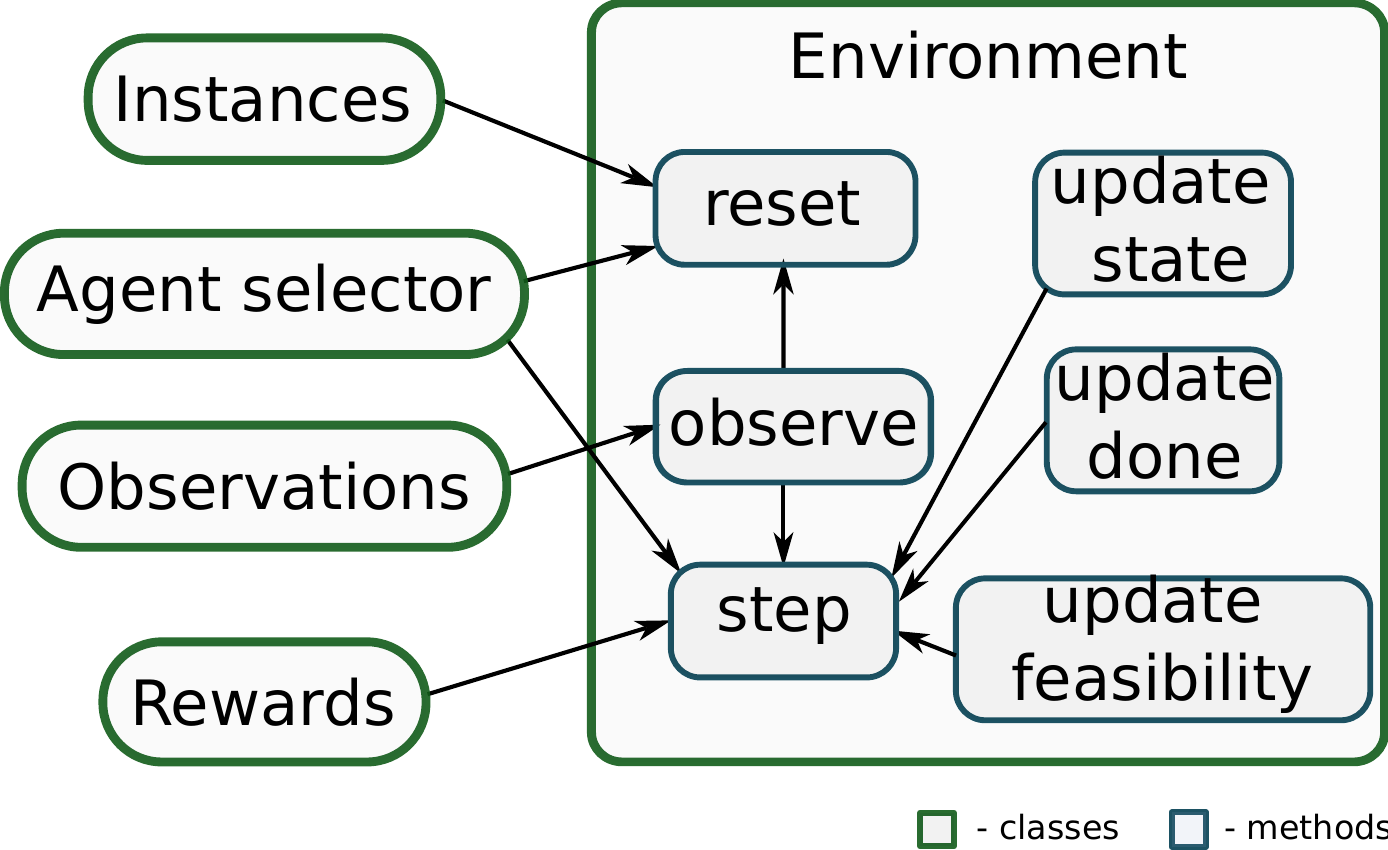}
\centering
\caption{Schematic representation of the library architecture.}
\label{env_fig}
\end{figure}

Each MAEnvs4VRP environment consists of four fundamental functional modules: Instance Generator, Observations Generator, Agent Selector, and Rewards classes (Fig. \ref{env_fig}). Each module operates autonomously, overseeing a crucial aspect of the environment's dynamics. This modular architecture, inspired by the principles of separation of concerns, simplifies experimentation, troubleshooting, and the expansion of individual components without compromising system stability. As a result, MAEnvs4VRP enables controlled exploration of aspects such as partial observability, reward shaping, and agent coordination strategies.

\textbf{Instance Generator Class.} The instance generation module defines the problem’s sample space and provides reproducible mechanisms for producing environment instances. It enables the rapid exploration of new instance distributions and the straightforward adoption of standard benchmark instance data commonly used to evaluate VRP solvers. This aims to narrow the gap between test beds for algorithm benchmarking used in RL and OR communities \cite{Accorsi20222}, allowing for a more objective performance comparison on common ground. 

Furthermore, this class includes the \texttt{augment\_generate\_instance} method, which enables batched instance augmentation by exploiting problem symmetries. Augmentation techniques have been widely used during training and inference to improve policy robustness and solution quality (e.g., \cite{kwon2020pomo, kim2022sym, LI2024124514}). In its current implementation, the method performs instance copying: it generates $N$ distinct problem instances and replicates each instance $K$ times, yielding a batched input of size $B = N \times K$. This copy-based augmentation facilitates multi-start rollouts on the same environment instance, as commonly employed in prior work (e.g., \cite{kwon2020pomo}). While the current implementation focuses on instance copying, the interface allows for straightforward extension to standard geometric augmentations, such as rotations or reflections of node coordinates (e.g., \cite{FITZPATRICK2024106787}).

\textbf{Observations class.} A critical aspect for successful agent training is their ability to retrieve valuable information from the environment to act on it. By decoupling observation computation from environment logic, the library facilitates research on feature engineering and exploration of the observation space, opening up the possibility of creating more aware and capable agents. 

Observations are organized in a hierarchical manner into \texttt{nodes\_static}, \texttt{nodes\_dynamic}, \texttt{agent}, \texttt{other\_agents}, and \texttt{global} features. This adaptable framework enables researchers to selectively expose information that directly influences the complexity of the learning task and the coordination strategy required (\cite{marl-book}).

\textbf{Agent Selector class.} Equivalent to \textit{PettingZoo}'s, the Agent Selector class manages the selection of the next agent that will interact with the environment, using the \texttt{\_next\_agent} method. This is an important aspect in multi-agent problems, particularly in time-dependent ones, as it enables the exploration of different strategies for choosing agents. For a broader class of VRP problems, particularly those involving time constraints or some form of randomness, more suitable agent selection functions, dependent on the state of the environment, can be designed or learned. 

At present, three distinct selection classes are available: \texttt{AgentSelector} (sequential, single-agent-like operation), \texttt{SmallestTimeAgentSelector} (real-time coordination (\cite{bono2020solving})), and \texttt{RandomSelector} (chooses an agent randomly among all active agents available). This explicit control over temporal coordination is a key differentiator of MAEnvs4VRP environments from single-agent setups. The sequence of agent selection, combined with accessible observations, is pivotal for scenarios involving inter-agent communication constraints. This is especially significant in dynamic and stochastic settings, where the strict preservation of causality and the prevention of future information leakage are fundamental requirements.

\textbf{Reward class.} Generally, the objective function to be optimized can vary significantly across the different VRP types, depending on the specific application. Additionally, even for the same VRP, the literature might present diverse objective functions. For instance, in the Capacitated Vehicle Routing Problem with Time Windows, conventional objectives might include minimizing the total route distance, reducing the total number of vehicles, or optimizing a linear combination of these factors. To easily account for this variability in the objective function's setup, the \texttt{Reward} class provides a straightforward reward design via the \texttt{get\_reward} method, enabling broader exploration of reward engineering. 

To account for violations of the problem constraints (e.g., maximum allowed tour duration, time-window violations, and unvisited customers), the method returns, for each environment step, a reward and a penalty (\cite{zhang2023first}). The ability to define a penalty, alongside and separately from the reward, is particularly advantageous for scenarios that aim at minimizing overall travel time or distance. This approach enables agents to decide whether to perform services or remain at the depot, without forcing a minimum number of services.

Currently, each environment provides access to both dense rewards (available at every step of the environment) and sparse rewards (available at the episode's conclusion). While dense rewards can provide the immediate feedback needed to stabilize early training, sparse rewards are often more representative of the primary objective, such as minimizing the total fleet distance or maximizing the number of served customers. This dual-reward structure allows for the customization of the learning signal to suit the specific requirements of different RL algorithms or optimization goals.

\subsection{Library Implementation and Interface}

Currently, the library includes 13 off-the-shelf environments. To provide representative examples of various variants, we implemented environments that account for both hard and soft time-window constraints, single- and multi-depot settings, deterministic/stochastic scenarios, and static/dynamic conditions. Currently, the library offers 13 off-the-shelf environments to simulate the following VRP: the Capacitated Vehicle Routing Problem with soft and hard time windows, Dynamic and Stochastic Vehicle Routing Problem with time windows (\cite{bono2020solving}), the Team Orienteering Problem, the Pickup and Delivery Problem, the Split Delivery Vehicle Routing Problem, the Prize-Collecting Vehicle Routing Problem, the Multi-depot Vehicle Routing Problem, and four implementations of multitask Vehicle Routing Problems (\cite{berto2024routefinder}). In line with \cite{Accorsi20222}, precise definitions of each environment, along with their observations and rewards, are provided in the library’s documentation.

\begin{figure}[ht]
\renewcommand{\figurename}{Code snippet}
\includegraphics[width=0.99\textwidth]{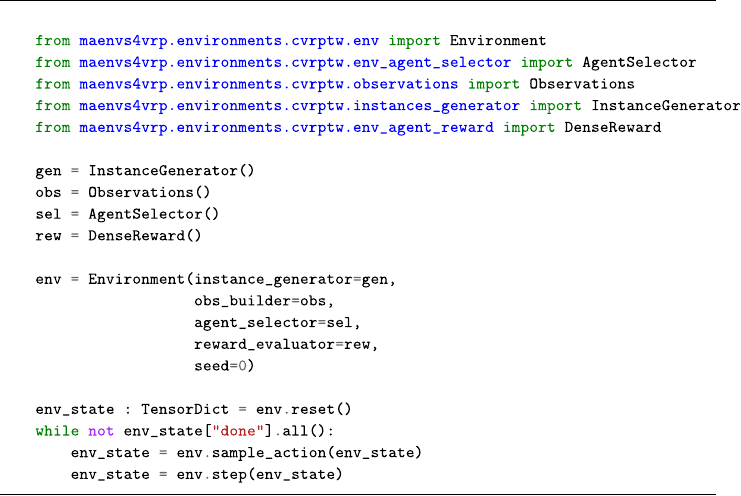}
\centering
\caption{Basic API usage example. Here, \texttt{env\_state} is a data container \texttt{TensorDict}. }
\label{lst:examp}
\end{figure}

All environments in MAEnvs4VRP adhere to a standardized API, ensuring consistency across problem variants. The library manages all data using TensorDicts, a dictionary-like structure introduced in TorchRL (\cite{bou2023torchrl}). This enables efficient tensor operations, streamlines data management, and facilitates the development of efficient batched (vectorized) environments that can run concurrently on GPUs. As illustrated in Code snippet \ref{lst:examp}, a typical episode run follows the logic found in most standard RL APIs. 

Following initialization, the environment advances through an iterative cycle of agent selection, observation gathering, and action determination, continuing until all vectorized environments are complete. Internally, an environment state \texttt{TensorDict} (\texttt{td\_state}) is maintained and updated throughout the rollout, storing all relevant information about nodes, agents, and environment dynamics. This sequential approach ensures that the environment state remains consistent with the agents' timeline during complex interactions. 

\section{Performance and Baselines}
\label{sec:experiments}

\begin{figure}[ht]
\includegraphics[width=0.95\textwidth]{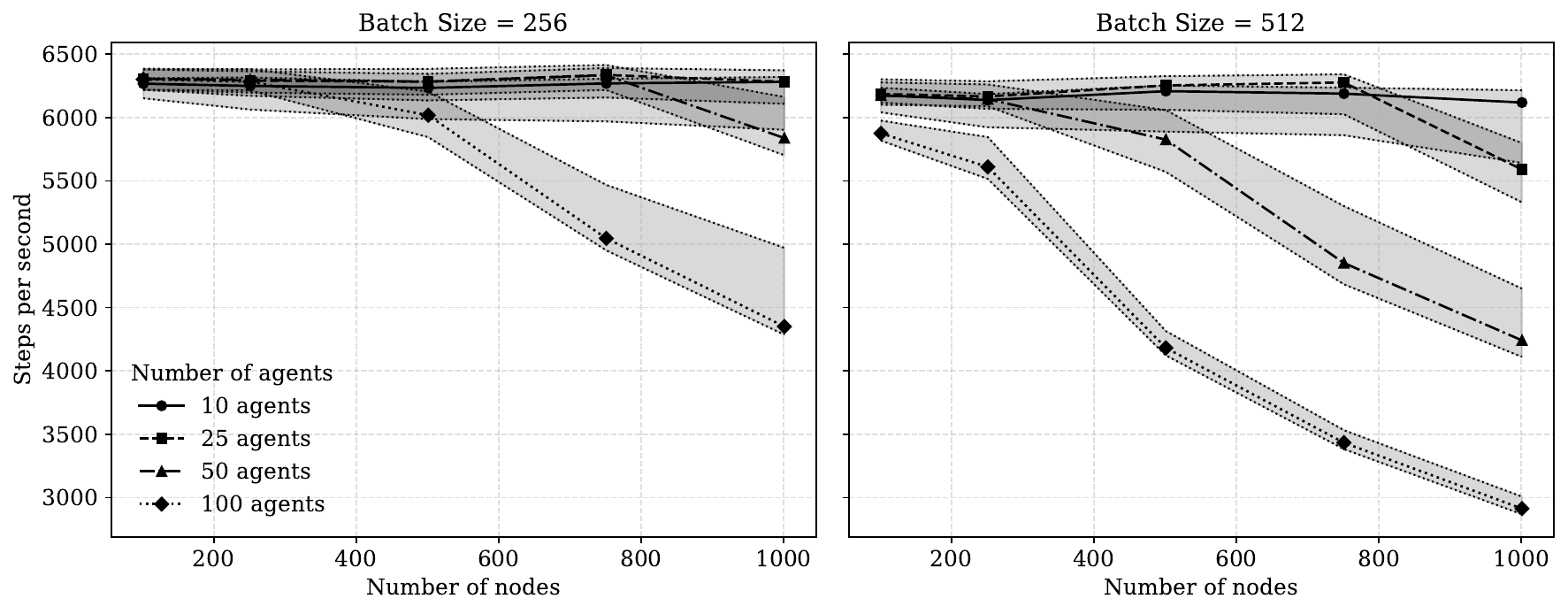}
\centering
\caption{Median and quartiles of the number of steps by number of nodes and number of agents over the environment rollouts using a random policy.}
\label{performance_fig}
\end{figure}

The computational performance of the environments, measured as the time required per simulation step, is influenced by several factors, including problem size, VRP dynamics, the policy used for service selection, specifically the neural network architecture underlying the decision process, and the hardware used (\cite{ Gymnasium2024}). We benchmark MAEnvs4VRP version 0.2.0, which is available as a static archive on the IJOC GitHub repository (\cite{ourpaper}). To assess baseline efficiency performance, we conducted 1,000 rollout experiments using a random policy, wherein agents select actions uniformly at random, across four sample problems: CVRPTW, DVRPTW, DSVRPTW, and MTVRP. Experiments were performed on an Nvidia GeForce RTX 4090 GPU.  We would like to emphasize that our primary objective in conducting these experiments is to validate MAEnvs4VRP, rather than to perform a comprehensive performance comparison across different problem configurations and hardware settings. Figure \ref{performance_fig} presents aggregated statistics from rollouts across the CVRPTW, DVRPTW, DSVRPTW, and MTVRP environments. The observed performance confirms that MAEnvs4VRP can throughput a large number of simulations in a computationally efficient manner, enabling rapid experimentation and training on accessible hardware configurations.

Additionally, we provide two baseline policy models, along with the training code.


\section{Discussion and future developments}
\label{sec:disc}

In this paper, we introduce MAEnvs4VRP, an open-source library for building and simulating multi-agent environments for vehicle routing problems. The library is built on a small number of core architectural principles, including sequential multi-agent decision-making, the adoption of \texttt{TensorDict} as the unified data container, and clear modular separation between problem functional components. These principles shape how components interact within the framework, but do not restrict the types of routing problems or operational settings that can be modeled.

The set of ready-to-use environments included in the library was implemented by instantiating these abstractions with problem-specific logic and constraints. These elements are contained within individual environment modules rather than hard-coded into the framework itself. This separation makes it easy to adapt existing environments, experiment with alternative modeling assumptions, or implement entirely new routing problems without changing the core architecture.

By combining this modular design with a user-friendly API and detailed documentation, MAEnvs4VRP aims to serve as both a practical research tool and a flexible foundation for future work in multi-agent combinatorial optimization. Alongside the library, we provide baseline training scripts to demonstrate its use and to support further exploration of reinforcement-learning approaches to multi-agent routing problems.

 \section*{Acknowledgements}

The first author gratefully acknowledges the Research Centre in Digital Services (CISeD), the Instituto Politécnico de Viseu, and the FCT - Foundation for Science and Technology, I.P., for their support during the work conducted under projects Refª UIDB/05583/2020 and 2023.13303.CPCA.A0.
 
\bibliography{bibfile.bib} 

@inproceedings{PettingZoo,
 author = {Terry, J and Black, Benjamin and Grammel, Nathaniel and Jayakumar, Mario  and Hari, Ananth  and Sullivan, Ryan and Santos, Luis S and Dieffendahl, Clemens and Horsch, Caroline and Perez-Vicente, Rodrigo and Williams, Niall  and Lokesh, Yashas  and Ravi , Praveen },
 booktitle = {Advances in Neural Information Processing Systems},
 editor = {M. Ranzato and A. Beygelzimer and Y. Dauphin and P.S. Liang and J. Wortman Vaughan},
 pages = {15032--15043},
 publisher = {Curran Associates, Inc.},
 title = {PettingZoo: Gym for Multi-Agent Reinforcement Learning},
 url = {https://proceedings.neurips.cc/paper_files/paper/2021/file/7ed2d3454c5eea71148b11d0c25104ff-Paper.pdf},
 volume = {34},
 year = {2021}
}

@misc{Gymnasium2024,
      title={Gymnasium: A Standard Interface for Reinforcement Learning Environments}, 
      author={Mark Towers and Ariel Kwiatkowski and Jordan Terry and John U. Balis and Gianluca De Cola and Tristan Deleu and Manuel Goulão and Andreas Kallinteris and Markus Krimmel and Arjun KG and Rodrigo Perez-Vicente and Andrea Pierré and Sander Schulhoff and Jun Jet Tai and Hannah Tan and Omar G. Younis},
      year={2024},
      eprint={2407.17032},
      archivePrefix={arXiv},
      primaryClass={cs.LG},
      url={https://arxiv.org/abs/2407.17032}, 
}

@article{Flatland,
  title={Flatland-RL : Multi-Agent Reinforcement Learning on Trains},
  author={Sharada Prasanna Mohanty and Erik Nygren and Florian Laurent and Manuel Schneider and Christian Vibe Scheller and Nilabha Bhattacharya and Jeremy Donald Watson and Adrian Egli and Christian Eichenberger and Christian Baumberger and Gereon Vienken and Irene Sturm and Guillaume Sartoretti and Giacomo Spigler},
  journal={ArXiv},
  year={2020},
  volume={abs/2012.05893}
}

@article{Accorsi20222,
title = {Guidelines for the computational testing of machine learning approaches to vehicle routing problems},
journal = {Operations Research Letters},
volume = {50},
number = {2},
pages = {229-234},
year = {2022},
issn = {0167-6377},
doi = {https://doi.org/10.1016/j.orl.2022.01.018},
url = {https://www.sciencedirect.com/science/article/pii/S0167637722000244},
author = {Luca Accorsi and Andrea Lodi and Daniele Vigo}
}

@article{brockman2016openai,
  title={Openai gym},
  author={Brockman, Greg and Cheung, Vicki and Pettersson, Ludwig and Schneider, Jonas and Schulman, John and Tang, Jie and Zaremba, Wojciech},
  journal={arXiv preprint arXiv:1606.01540},
  year={2016}
}

@article{Bello2016,
archivePrefix = {arXiv},
arxivId = {1611.09940},
author = {Bello, Irwan and Pham, Hieu and Le, Quoc V. and Norouzi, Mohammad and Bengio, Samy},
doi = {10.1146/annurev.cellbio.15.1.81},
eprint = {1611.09940},
title = {{Neural Combinatorial Optimization with Reinforcement Learning}},
journal={Proceedings of the 5th International Conference on Learning Representations (ICLR)},
year={2017},
url = {http://arxiv.org/abs/1611.09940},
}

@incollection{Vinyals2015,
title = {Pointer Networks},
author = {Vinyals, Oriol and Fortunato, Meire and Jaitly, Navdeep},
booktitle = {Advances in Neural Information Processing Systems 28},
editor = {C. Cortes and N. D. Lawrence and D. D. Lee and M. Sugiyama and R. Garnett},
pages = {2692--2700},
year = {2015},
publisher = {Curran Associates, Inc.},
url = {http://papers.nips.cc/paper/5866-pointer-networks.pdf}
}

@inproceedings{Nazari2018,
archivePrefix = {arXiv},
arxivId = {1802.04240},
author = {Nazari, Mohammadreza and Oroojlooy, Afshin and Snyder, Lawrence V. and Tak{\'{a}}{\v{c}}, Martin},
title = {{Deep Reinforcement Learning for Solving the Vehicle Routing Problem}},
booktitle={Proceedings Neural Information Processing Systems (NIPS)},
pages = {9839-9849},
url = {http://arxiv.org/abs/1802.04240},
year = {2018}
}

@article{Kool2019a,
archivePrefix = {arXiv},
arxivId = {1803.08475},
author = {Kool, Wouter and {Van Hoof}, Herke and Welling, Max},
eprint = {1803.08475},
journal = {7th International Conference on Learning Representations, ICLR 2019},
pages = {1--25},
title = {{Attention, learn to solve routing problems!}},
year = {2019}
}

@article{MAZYAVKINA2021105400,
title = {Reinforcement learning for combinatorial optimization: A survey},
journal = {Computers \& Operations Research},
volume = {134},
pages = {105400},
year = {2021},
issn = {0305-0548},
doi = {https://doi.org/10.1016/j.cor.2021.105400},
url = {https://www.sciencedirect.com/science/article/pii/S0305054821001660},
author = {Nina Mazyavkina and Sergey Sviridov and Sergei Ivanov and Evgeny Burnaev}
}

@misc{hubbs2020orgym,
      title={OR-Gym: A Reinforcement Learning Library for Operations Research Problems}, 
      author={Christian D. Hubbs and Hector D. Perez and Owais Sarwar and Nikolaos V. Sahinidis and Ignacio E. Grossmann and John M. Wassick},
      year={2020},
      eprint={2008.06319},
      archivePrefix={arXiv},
      primaryClass={cs.AI}
}

@misc{bonnet2023jumanji,
    title={Jumanji: a Diverse Suite of Scalable Reinforcement Learning Environments in JAX},
    author={
        Clément Bonnet and Daniel Luo and Donal Byrne and Shikha Surana and Vincent Coyette and
        Paul Duckworth and Laurence I. Midgley and Tristan Kalloniatis and Sasha Abramowitz and
        Cemlyn N. Waters and Andries P. Smit and Nathan Grinsztajn and Ulrich A. Mbou Sob and
        Omayma Mahjoub and Elshadai Tegegn and Mohamed A. Mimouni and Raphael Boige and
        Ruan de Kock and Daniel Furelos-Blanco and Victor Le and Arnu Pretorius and
        Alexandre Laterre
    },
    year={2023},
    eprint={2306.09884},
    url={https://arxiv.org/abs/2306.09884},
    archivePrefix={arXiv},
    primaryClass={cs.LG}
}

@article{berto2023rl4co,
    title = {{RL4CO}: an Extensive Reinforcement Learning for Combinatorial Optimization Benchmark},
    author={Federico Berto and Chuanbo Hua and Junyoung Park and Minsu Kim and Hyeonah Kim and Jiwoo Son and Haeyeon Kim and Joungho Kim and Jinkyoo Park},
    journal={arXiv preprint arXiv:2306.17100},
    year={2023},
    url = {https://github.com/kaist-silab/rl4co}
}

@misc{wan2023rlor,
      title={RLOR: A Flexible Framework of Deep Reinforcement Learning for Operation Research}, 
      author={Ching Pui Wan and Tung Li and Jason Min Wang},
      year={2023},
      eprint={2303.13117},
      archivePrefix={arXiv},
      primaryClass={math.OC}
}

@ARTICLE{Lireview2022,
  author={Li, Bingjie and Wu, Guohua and He, Yongming and Fan, Mingfeng and Pedrycz, Witold},
  journal={IEEE/CAA Journal of Automatica Sinica}, 
  title={An Overview and Experimental Study of Learning-Based Optimization Algorithms for the Vehicle Routing Problem}, 
  year={2022},
  volume={9},
  number={7},
  pages={1115-1138},
  doi={10.1109/JAS.2022.105677}}

@article{SHI2023773,
title = {A Brief Survey on Learning Based Methods for Vehicle Routing Problems},
journal = {Procedia Computer Science},
volume = {221},
pages = {773-780},
year = {2023},
note = {Tenth International Conference on Information Technology and Quantitative Management (ITQM 2023)},
issn = {1877-0509},
doi = {https://doi.org/10.1016/j.procs.2023.08.050},
url = {https://www.sciencedirect.com/science/article/pii/S1877050923008086},
author = {Ruiyang Shi and Lingfeng Niu}
}

@ARTICLE{fuertes2023,
  author={Fuertes, Daniel and del-Blanco, Carlos R. and Jaureguizar, Fernando and García, Narciso},
  journal={IEEE Transactions on Intelligent Transportation Systems}, 
  title={TOP-Former: A Multi-Agent Transformer Approach for the Team Orienteering Problem}, 
  year={2025},
  volume={26},
  number={9},
  pages={13799-13810},
  doi={10.1109/TITS.2025.3566157}}

@article{kwon2020pomo,
  title={Pomo: Policy optimization with multiple optima for reinforcement learning},
  author={Kwon, Yeong-Dae and Choo, Jinho and Kim, Byoungjip and Yoon, Iljoo and Gwon, Youngjune and Min, Seungjai},
  journal={Advances in Neural Information Processing Systems},
  volume={33},
  pages={21188--21198},
  year={2020}
}

@article{arishi2023multi,
  title={A multi-agent deep reinforcement learning approach for solving the multi-depot vehicle routing problem},
  author={Arishi, Ali and Krishnan, Krishna},
  journal={Journal of Management Analytics},
  volume={10},
  number={3},
  pages={493--515},
  year={2023},
  publisher={Taylor \& Francis}
}

@article{bono2020solving,
  title={Solving multi-agent routing problems using deep attention mechanisms},
  author={Bono, Guillaume and Dibangoye, Jilles S and Simonin, Olivier and Matignon, La{\"e}titia and Pereyron, Florian},
  journal={IEEE Transactions on Intelligent Transportation Systems},
  volume={22},
  number={12},
  pages={7804--7813},
  year={2020},
  publisher={IEEE}
}

@article{zhang2020multi,
  title={Multi-vehicle routing problems with soft time windows: A multi-agent reinforcement learning approach},
  author={Zhang, Ke and He, Fang and Zhang, Zhengchao and Lin, Xi and Li, Meng},
  journal={Transportation Research Part C: Emerging Technologies},
  volume={121},
  pages={102861},
  year={2020},
  publisher={Elsevier}
}

@article{wu2024neural,
  title={Neural Combinatorial Optimization Algorithms for Solving Vehicle Routing Problems: A Comprehensive Survey with Perspectives},
  author={Wu, Xuan and Wang, Di and Wen, Lijie and Xiao, Yubin and Wu, Chunguo and Wu, Yuesong and Yu, Chaoyu and Maskell, Douglas L and Zhou, You},
  journal={arXiv preprint arXiv:2406.00415},
  year={2024}
}

@article{liu2024multi,
title={Multi-Task Learning for Routing Problem with Cross-Problem Zero-Shot Generalization},
author={Liu, Fei and Lin, Xi and Zhang, Qingfu and Tong, Xialiang and Yuan, Mingxuan},
journal={arXiv preprint arXiv:2402.16891},
year={2024}
}

@inproceedings{zong2022mapdp,
  title={Mapdp: Cooperative multi-agent reinforcement learning to solve pickup and delivery problems},
  author={Zong, Zefang and Zheng, Meng and Li, Yong and Jin, Depeng},
  booktitle={Proceedings of the AAAI Conference on Artificial Intelligence},
  volume={36},
  number={9},
  pages={9980--9988},
  year={2022}
}

@article{LI2024124514,
title = {Solving pick-up and delivery problems via deep reinforcement learning based symmetric neural optimization},
journal = {Expert Systems with Applications},
volume = {255},
pages = {124514},
year = {2024},
issn = {0957-4174},
doi = {https://doi.org/10.1016/j.eswa.2024.124514},
url = {https://www.sciencedirect.com/science/article/pii/S0957417424013812},
author = {Jinqi Li and Yunyun Niu and Guodong Zhu and Jianhua Xiao},
}

@misc{bou2023torchrl,
      title={TorchRL: A data-driven decision-making library for PyTorch},
      author={Albert Bou and Matteo Bettini and Sebastian Dittert and Vikash Kumar and Shagun Sodhani and Xiaomeng Yang and Gianni De Fabritiis and Vincent Moens},
      year={2023},
      eprint={2306.00577},
      archivePrefix={arXiv},
      primaryClass={cs.LG}
}

@inbook{pytorch,
author = {Paszke, Adam and Gross, Sam and Massa, Francisco and Lerer, Adam and Bradbury, James and Chanan, Gregory and Killeen, Trevor and Lin, Zeming and Gimelshein, Natalia and Antiga, Luca and Desmaison, Alban and K\"{o}pf, Andreas and Yang, Edward and DeVito, Zach and Raison, Martin and Tejani, Alykhan and Chilamkurthy, Sasank and Steiner, Benoit and Fang, Lu and Bai, Junjie and Chintala, Soumith},
title = {PyTorch: an imperative style, high-performance deep learning library},
year = {2019},
publisher = {Curran Associates Inc.},
address = {Red Hook, NY, USA},
booktitle = {Proceedings of the 33rd International Conference on Neural Information Processing Systems},
articleno = {721},
numpages = {12}
}

@article{zhou2024mvmoe,
  title={MVMoE: Multi-Task Vehicle Routing Solver with Mixture-of-Experts},
  author={Zhou, Jianan and Cao, Zhiguang and Wu, Yaoxin and Song, Wen and Ma, Yining and Zhang, Jie and Xu, Chi},
  journal={arXiv preprint arXiv:2405.01029},
  year={2024}
}

@article{berto2024routefinder,
  title={RouteFinder: Towards foundation models for vehicle routing problems},
  author={Berto, Federico and Hua, Chuanbo and Zepeda, Nayeli Gast and Hottung, Andr{\'e} and Wouda, Niels and Lan, Leon and Tierney, Kevin and Park, Jinkyoo},
  journal={arXiv preprint arXiv:2406.15007},
  year={2024}
}

@article{zhang2023coordinated,
  title={Coordinated multi-agent hierarchical deep reinforcement learning to solve multi-trip vehicle routing problems with soft time windows},
  author={Zhang, Zixian and Qi, Geqi and Guan, Wei},
  journal={IET Intelligent Transport Systems},
  volume={17},
  number={10},
  pages={2034--2051},
  year={2023},
  publisher={Wiley Online Library}
}

@ARTICLE{10417723,
  author={Xiang, Chuankai and Wu, Zhibin and Tu, Jiancheng and Huang, Jun},
  journal={IEEE Transactions on Intelligent Transportation Systems}, 
  title={Centralized Deep Reinforcement Learning Method for Dynamic Multi-Vehicle Pickup and Delivery Problem With Crowdshippers}, 
  year={2024},
  volume={25},
  number={8},
  pages={9253-9267},
  doi={10.1109/TITS.2024.3352143}}

@article{hu2023marllib,
  title={Marllib: A scalable and efficient multi-agent reinforcement learning library},
  author={Hu, Siyi and Zhong, Yifan and Gao, Minquan and Wang, Weixun and Dong, Hao and Liang, Xiaodan and Li, Zhihui and Chang, Xiaojun and Yang, Yaodong},
  journal={Journal of Machine Learning Research},
  volume={24},
  number={315},
  pages={1--23},
  year={2023}
}

@article{raffin2021stable,
  title={Stable-baselines3: Reliable reinforcement learning implementations},
  author={Raffin, Antonin and Hill, Ashley and Gleave, Adam and Kanervisto, Anssi and Ernestus, Maximilian and Dormann, Noah},
  journal={Journal of Machine Learning Research},
  volume={22},
  number={268},
  pages={1--8},
  year={2021}
}

@article{bettini2024benchmarl,
  title={Benchmarl: Benchmarking multi-agent reinforcement learning},
  author={Bettini, Matteo and Prorok, Amanda and Moens, Vincent},
  journal={Journal of Machine Learning Research},
  volume={25},
  number={217},
  pages={1--10},
  year={2024}
}

@article{zhou2024learning,
  title={Learning-Based Optimization Algorithms for Routing Problems: Bibliometric Analysis and Literature Review},
  author={Zhou, Guanghui and Li, Xiaoyi and Li, Dengyuhui and Bian, Junsong},
  journal={IEEE Transactions on Intelligent Transportation Systems},
  year={2024},
  publisher={IEEE}
}

@article{GUO2023103095,
title = {Deep attention models with dimension-reduction and gate mechanisms for solving practical time-dependent vehicle routing problems},
journal = {Transportation Research Part E: Logistics and Transportation Review},
volume = {173},
pages = {103095},
year = {2023},
issn = {1366-5545},
doi = {https://doi.org/10.1016/j.tre.2023.103095},
url = {https://www.sciencedirect.com/science/article/pii/S1366554523000832},
author = {Feng Guo and Qu Wei and Miao Wang and Zhaoxia Guo and Stein W. Wallace}}

@article{pan2023deep,
  title={Deep reinforcement learning for the dynamic and uncertain vehicle routing problem},
  author={Pan, Weixu and Liu, Shi Qiang},
  journal={Applied Intelligence},
  volume={53},
  number={1},
  pages={405--422},
  year={2023},
  publisher={Springer}
}

@book{shoham2008multiagent,
  title={Multiagent systems: Algorithmic, game-theoretic, and logical foundations},
  author={Shoham, Yoav and Leyton-Brown, Kevin},
  year={2008},
  publisher={Cambridge University Press}
}

@article{balaji2019orl,
  title={Orl: Reinforcement learning benchmarks for online stochastic optimization problems},
  author={Balaji, Bharathan and Bell-Masterson, Jordan and Bilgin, Enes and Damianou, Andreas and Garcia, Pablo Moreno and Jain, Arpit and Luo, Runfei and Maggiar, Alvaro and Narayanaswamy, Balakrishnan and Ye, Chun},
  journal={arXiv preprint arXiv:1911.10641},
  year={2019}
}

@article{biagioni2022graphenv,
  title={graphenv: a Python library for reinforcement learning on graph search spaces},
  author={Biagioni, David and Tripp, Charles Edison and Clark, Struan and Duplyakin, Dmitry and Law, Jeffrey and John, Peter C St},
  journal={Journal of Open Source Software},
  volume={7},
  number={77},
  pages={4621},
  year={2022}
}

@misc{berto2024parco,
      title={Parallel AutoRegressive Models for Multi-Agent Combinatorial Optimization}, 
      author={Federico Berto and Chuanbo Hua and Laurin Luttmann and Jiwoo Son and Junyoung Park and Kyuree Ahn and Changhyun Kwon and Lin Xie and Jinkyoo Park},
      year={2025},
      eprint={2409.03811},
      archivePrefix={arXiv},
      primaryClass={cs.MA},
      url={https://arxiv.org/abs/2409.03811}, 
}

@article{zhang2023first,
  title={The first AI4TSP competition: Learning to solve stochastic routing problems},
  author={Zhang, Yingqian and Bliek, Laurens and da Costa, Paulo and Afshar, Reza Refaei and Reijnen, Robbert and Catshoek, Tom and Vos, Dani{\"e}l and Verwer, Sicco and Schmitt-Ulms, Fynn and Hottung, Andr{\'e} and others},
  journal={Artificial Intelligence},
  volume={319},
  pages={103918},
  year={2023},
  publisher={Elsevier}
}

@article{kim2022sym,
  title={Sym-nco: Leveraging symmetricity for neural combinatorial optimization},
  author={Kim, Minsu and Park, Junyoung and Park, Jinkyoo},
  journal={Advances in Neural Information Processing Systems},
  volume={35},
  pages={1936--1949},
  year={2022}
}

@article{thyssens2023routing,
  title={Routing Arena: A Benchmark Suite for Neural Routing Solvers},
  author={Thyssens, Daniela and Dernedde, Tim and Falkner, Jonas K and Schmidt-Thieme, Lars},
  journal={arXiv preprint arXiv:2310.04140},
  year={2023}
}

@inproceedings{liu20242d,
  title={2D-Ptr: 2D Array Pointer Network for Solving the Heterogeneous Capacitated Vehicle Routing Problem},
  author={Liu, Qidong and Liu, Chaoyue and Niu, Shaoyao and Long, Cheng and Zhang, Jie and Xu, Mingliang},
  booktitle={Proceedings of the 23rd International Conference on Autonomous Agents and Multiagent Systems},
  pages={1238--1246},
  year={2024}
}

@book{ marl-book,
  author = {Stefano V. Albrecht and Filippos Christianos and Lukas Sch\"afer},
  title = {Multi-Agent Reinforcement Learning: Foundations and Modern Approaches},
  publisher = {MIT Press},
  year = {2024},
  url = {https://www.marl-book.com}
}

@article{menda2018deep,
  title={Deep reinforcement learning for event-driven multi-agent decision processes},
  author={Menda, Kunal and Chen, Yi-Chun and Grana, Justin and Bono, James W and Tracey, Brendan D and Kochenderfer, Mykel J and Wolpert, David},
  journal={IEEE Transactions on Intelligent Transportation Systems},
  volume={20},
  number={4},
  pages={1259--1268},
  year={2018},
  publisher={IEEE}
}

@article{braekers2016vehicle,
  title={The vehicle routing problem: State of the art classification and review},
  author={Braekers, Kris and Ramaekers, Katrien and Van Nieuwenhuyse, Inneke},
  journal={Computers \& industrial engineering},
  volume={99},
  pages={300--313},
  year={2016},
  publisher={Elsevier}
}

@article{FITZPATRICK2024106787,
title = {A scalable learning approach for the capacitated vehicle routing problem},
journal = {Computers \& Operations Research},
volume = {171},
pages = {106787},
year = {2024},
issn = {0305-0548},
doi = {https://doi.org/10.1016/j.cor.2024.106787},
url = {https://www.sciencedirect.com/science/article/pii/S0305054824002594},
author = {James Fitzpatrick and Deepak Ajwani and Paula Carroll}
}

@misc{ourpaper,
  author =     {Gama, Ricardo and Fernandes, Hugo and Fuertes, Daniel and del-Blanco, Carlos R. and Cunha, Ricardo},
  publisher =  {INFORMS Journal on Computing},
  title =      {Multi-Agent Environments for Vehicle Routing Problems},
  year =       {2026},
  doi =        {10.1287/ijoc.2025.1211.cd},
  note =       {Available for download at https://github.com/INFORMSJoC/2025.1211},
}


\end{document}